\documentclass[11pt]{article}

\usepackage[preprint]{acl}

\usepackage{times}
\usepackage{latexsym}
\usepackage{booktabs}
\usepackage{multirow}
\usepackage{pifont}

\usepackage{pifont}
\usepackage[table]{xcolor}

\newcommand{\cmark}{\textcolor{green!60!black}{\ding{51}}}
\newcommand{\xmark}{\textcolor{red!70!black}{\ding{55}}}

\usepackage[T1]{fontenc}

\usepackage[utf8]{inputenc}

\usepackage{microtype}

\usepackage{inconsolata}

\usepackage{graphicx}

\title{The Atomic Instruction Gap: Instruction-Tuned LLMs Struggle with Simple, Self-Contained Directives}

\author{Henry Lim \\
  Singapore University of Technology  \\
  and Design, Singapore \\
  \texttt{henry\_lim@mymail.sutd.edu.sg} \\\And
  Kwan Hui Lim \\
  Singapore University of Technology \\
  and Design, Singapore \\
  \texttt{kwanhui\_lim@sutd.edu.sg} \\}

\begin{document}
\maketitle
\begin{abstract}
Instruction-tuned large language models (IT-LLMs) exhibit strong zero-shot reasoning, yet their ability to execute simple, self-contained instructions remains underexplored, despite this being foundational to complex instruction-following. We evaluate 20 IT-LLMs on modified MMLU and MMLU-Pro benchmarks, by systematically varying the format of option labels (alphabetic, numeric, Roman) while keeping their meaning identical under four paradigms, namely: 
(1) With explicit instructions, label changes cause large performance shifts (e.g., -30.45\% for Roman vs. numeric), revealing instruction-format bias. (2) Without instructions, performance drops further (up to -10.84\%) and label sensitivity intensifies, underscoring the role of explicit guidance. (3) When option contents are removed, models fail random-choice baselines except with numeric labels, suggesting weak adherence to atomic directives. (4) Three-shot exemplars yield no significant gains in robustness or fidelity, and generation analyses show persistent label errors, especially for non-numeric formats. 
Across model sizes, larger LLMs achieve higher accuracy but remain inconsistent in instruction adherence. These results expose the insufficiencies of current instruction-tuning paradigms and highlight the need for evaluation methods and training strategies that explicitly target atomic instruction-following.
\end{abstract}

\section{Introduction}

\begin{figure*}[t]
  \includegraphics[width=\textwidth]{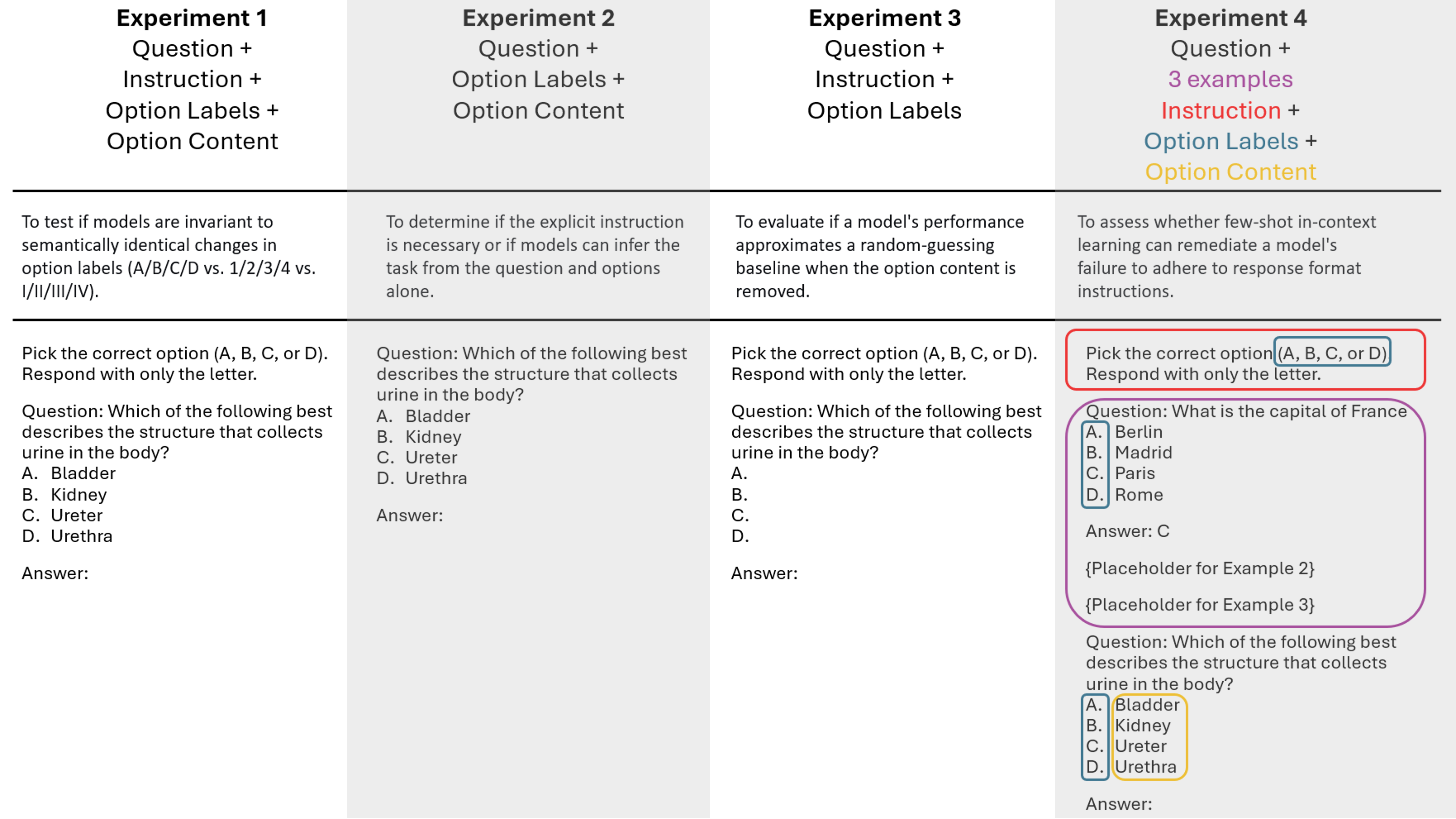}
  \caption{\textbf{Overview of experimental setups.} Each experiment systematically manipulates components of multiple-choice question prompts - instructions, option labels, option content, and few-shot examples—to test how model performance and response adherence vary under different input configurations. Experiment 1 includes all components; Experiment 2 removes the explicit instruction to test whether models can infer the task; Experiment 3 removes option content to approximate a random-guessing baseline; and Experiment 4 adds few-shot examples to evaluate whether in-context learning mitigates instruction-following failures. All experiments are repeated across three sets of option labels: alphabetical (A/B/C/D), numerical (1/2/3/4), and Roman (I/II/III/IV).}
  \label{fig:experiments}
\end{figure*}

Instruction tuning has significantly improved the 0-shot capabilities of large language models (LLMs), enabling complex task execution via natural language prompts \cite{wei2022finetunedlanguagemodelszeroshot, iyer2023optimlscalinglanguagemodel, sanh2022multitaskpromptedtrainingenables, ouyang2022traininglanguagemodelsfollow}. However, while prior work focuses on multi-step \cite{jiang2024followbenchmultilevelfinegrainedconstraints, chen2024sifobenchmarkinvestigatingsequential}, or uncommon instructions \cite{li2024instructionfollowingevaluationverbalizermanipulation}, the ability of instruction-tuned LLMs (IT-LLMs) to follow atomic instructions - simple, unambiguous directives - remains critically underexplored. This gap is consequential: atomic adherence underpins reliable performance in deterministic applications and compositional reasoning \cite{hayati2025chainofinstructionscompositionalinstructiontuning, he2024complexsimpleenhancingmulticonstraint, sun2024coniferimprovingcomplexconstrained}.

We empirically demonstrate that IT-LLMs exhibit profound fragility to semantically invariant atomic instructions, using MMLU \cite{hendrycks2021measuringmassivemultitasklanguage} and MMLU-Pro \cite{wang2024mmluprorobustchallengingmultitask}. Through four controlled experiments, we reveal systematic failures. Our work is motivated by two key factors: (1) Real-world relevance - Practical applications often involve tasks coupled with instructions, yet the impact of instruction-following ability on task performance remains unclear. Hence, we pick two knowledge-intensive benchmarks. (2) Evaluation practicality - MCQs provide a straightforward format for automated evaluation and have been adopted for assessing LLM capabilities \cite{robinson2023leveraginglargelanguagemodels}.

\textbf{Experiment 1} varies option labels (alphabetic, numeric, Roman) with explicit instructions. Despite semantic equivalence, models show drastic performance gaps (e.g., 30.44\% drop for Roman vs. numeric labels on MMLU), exposing instruction bias to surface formats. \textbf{Experiment 2} removes instructions entirely. Performance degrades significantly (e.g., -10.84\% for Roman labels on MMLU), confirming instructions' necessity. \textbf{Experiment 3} deprives options of content, retaining only labels and instructions. Models approach chance only for numeric labels, collapsing otherwise (alphabetic: 13.90\%, Roman: 12.97\% vs. 25\% baseline on MMLU), proving inherent inability to execute atomic directives. Experiment 4 investigates whether 3-shot in-context learning can mitigate these failures by providing explicit format demonstrations. Finally, we conduct a fine-grained Analysis of Generated Answers to measure the proportion of syntactically valid outputs, revealing that models frequently fail to produce responses within the instructed label set, even with examples.


Our contributions are fourfold: (1) We expose IT-LLMs' fundamental failure to follow atomic instructions, showing that semantically irrelevant label-style variations cause statistically significant performance drops. (2) We demonstrate that this fragility is not solved by standard interventions, persisting despite model scaling, iterative instruction-tuning, and few-shot in-context learning. (3) We trace the origin of this symbolic bias to the base models, indicating it is a pre-training artifact that instruction-tuning fails to overwrite. (4) We show this failure is fundamental, as models cannot execute the instruction when deprived of task content, often performing below random chance.

\section{Related Work}

\textbf{Instruction-Following Evaluation} Recent work has established benchmarks to assess the generality and robustness of IT-LLMs. Frameworks like Super-NaturalInstructions \cite{wang2022supernaturalinstructionsgeneralizationdeclarativeinstructions} and AlpacaEval \cite{dubois2025lengthcontrolledalpacaevalsimpleway} evaluate models on diverse, complex tasks, while InstructEval IFEval \cite{zhou2023instructionfollowingevaluationlargelanguage} focus on fine-grained constraints adherence to constraints. However, these benchmarks primarily test compositional or multi-step instructions \cite{li2025structflowbenchstructuredflowbenchmark, chen2024sifobenchmarkinvestigatingsequential}, leaving a gap in understanding how models handle atomic instructions, despite their foundational role in more complex behavior \cite{hayati2025chainofinstructionscompositionalinstructiontuning}. Our work bridges this gap by decoupling task semantics from instruction adherence, revealing that even trivial variations (i.e., label shifts) disrupt performance. Additionally, our work facilitates easy extension to existing tasks, such as Classification, with minimal human efforts. For instance, whereas \citet{zhou2023instructionfollowingevaluationlargelanguage} manually constructed approximately 500 prompts, our evaluation leverages around 14,000 and 12,000 samples from MMLU and MMLU-Pro, respectively.

\textbf{Prompt Sensitivity and Robustness} The fragility of LLMs to prompt phrasing is well-documented: \cite{mishra2022reframinginstructionalpromptsgptks} show that minor rewrites (e.g., passive vs. active voice) significantly alter outputs, while \cite{turpin2023languagemodelsdontsay} demonstrate that models frequently ignore implicit constraints. Similar issues arise in multiple-choice settings, where models exhibit bias toward specific option positions or formats \cite{robinson2023leveraginglargelanguagemodels, wang2023largelanguagemodelszeroshot}. However, prior work focuses on task-driven sensitivity (e.g., accuracy drops due to misleading distractors), whereas we isolate instruction-driven sensitivity (e.g., semantically neutral changes to labels or phrasing). Our findings align with \cite{mckenzie2024inversescalingbiggerisnt}, who observe that larger models may amplify spurious prompt correlations, but we further show that even iterative improvements (e.g., Qwen’s versions) do not guarantee robustness to atomic instruction variations.
Despite these advances, recent work highlights persistent weaknesses such as being sensitive to prompt phrasing \cite {mishra2022reframinginstructionalpromptsgptks} and overfitting to templated instructions \cite{zhou2023largelanguagemodelshumanlevel, mckenzie2024inversescalingbiggerisnt, shah2022goalmisgeneralizationcorrectspecifications}. \cite{turpin2023languagemodelsdontsay} further show that LLMs often ignore implicit constraints. Unlike prior work, which focuses on complex or ambiguous instructions, we identify systematic failures in simple and atomic instructions, suggesting fundamental gaps in instruction comprehension beyond instruction-tuning.

\section{Methodology}

\begin{table*}[t]
  \centering
  \resizebox{\textwidth}{!}{%
    \begin{tabular}{l*{22}{c}}
      \toprule
      \multirow{2}{*}{Option Labels} & \multicolumn{7}{c}{QWEN} & \multicolumn{6}{c}{LLAMA} & \multicolumn{3}{c}{MISTRAL} & \multicolumn{2}{c}{PHI 3} & \multicolumn{2}{c}{OLMo-2} & \multicolumn{2}{c}{STATS}\\
      \cmidrule(lr){2-8} \cmidrule(lr){9-14} \cmidrule(lr){15-17} \cmidrule(lr){18-19} \cmidrule(lr){20-21} \cmidrule(lr){22-23}
       & 0.5B & 1.5B & 3B & 7B & 14B & 32B & 72B & 1B & 3B & 8B & 8B-M & 70B & 70B & 7B & 7B & 7B & 3.8B & 14B & 1B & 32B & Avg & F\\
      \midrule
      A / B / C / D & 5.47 & 17.86 & 13.12 & 70.35 & 76.18 & 79.91 & 79.96 & 44.70 & 55.92 & 29.10 & 57.92 & 47.48 & 31.01 & 25.65 & 43.03 & 56.63 & 0.03 & 0.00 & 37.85 & 71.12 & 42.16 & 5\\
      1 / 2 / 3 / 4 & 37.73 & 56.69 & 62.68 & 70.80 & 75.86 & 79.57 & 84.06 & 44.65 & 57.72 & 65.38 & 62.80 & 80.60 & 79.54 & 49.55 & 56.54 & 58.62 & 64.37 & 63.56 & 38.28 & 72.28 & 63.06 & 0\\
      I / II / III / IV & 0.36 & 0.78 & 21.25 & 67.94 & 68.42 & 78.72 & 76.25 & 39.50 & 53.72 & 2.93 & 37.15 & 8.40 & 15.02 & 8.44 & 24.73 & 45.38 & 0.00 & 0.00 & 31.83 & 71.55 & 32.62 & 10\\
      Min-Max Range & 37.37 & 55.91 & 49.56 & 2.86 & 7.76 & 1.19 & 7.81 & 5.20 & 4.00 & 62.45 & 25.65 & 72.20 & 64.52 & 41.11 & 31.81 & 13.24 & 64.37 & 63.56 & 6.45 & 1.16 & 30.45 & 10\\
      \midrule
      A / B / C / D  & 0.97 & 4.57 & 3.94 & 40.81 & 52.52 & 55.81 & 49.05 & 19.56 & 31.65 & 8.37 & 30.46 & 25.28 & 15.04 & 11.19 & 14.50 & 26.96 & 0.18 & 30.45 & 12.29 & 43.69 & 22.33 & 6\\
      1 / 2 / 3 / 4 & 13.57 & 25.41 & 31.56 & 40.95 & 48.76 & 54.12 & 50.10 & 19.06 & 32.53 & 39.37 & 35.26 & 53.09 & 52.37 & 22.65 & 26.26 & 29.72 & 5.17 & 37.13 & 14.37 & 44.87 & 35.50 & 0\\
      I / II / III / IV & 0.04 & 0.94 & 8.86 & 39.45 & 44.79 & 52.68 & 43.54 & 13.84 & 28.0 & 2.79 & 20.61 & 8.06 & 53.09 & 1.03 & 9.36 & 22.45 & 0.06 & 23.30 & 11.83 & 42.13 & 20.17 & 9\\
      Min-Max Range & 13.53 & 24.47 & 27.62 & 1.50 & 7.73 & 3.13 & 6.56 & 5.72 & 4.53 & 36.58 & 14.65 & 45.03 & 38.05 & 21.62 & 16.90 & 7.27 & 34.48 & 41.52 & 2.54 & 2.74 & 15.33 & 9\\
      \bottomrule
    \end{tabular}%
  }

  \caption{Accuracy (\%) on MMLU (top) and MMLU-Pro (bottom). 
Min-Max Range: performance spread across formats. 
F: failures (below random baseline: 25\% MMLU, 10\% MMLU-Pro)}
  \label{tab:main_results}
\end{table*}

\begin{table*}[t]
  \centering
  \resizebox{\textwidth}{!}{%
    \begin{tabular}{l*{19}{c}}
      \toprule
      \multirow{2}{*}{Option Labels} & \multicolumn{6}{c}{QWEN} & \multicolumn{4}{c}{LLAMA} & \multicolumn{3}{c}{MISTRAL} & \multicolumn{2}{c}{PHI 3} & \multicolumn{2}{c}{OLMo-2} & \multicolumn{2}{c}{STATS}\\
      \cmidrule(lr){2-7} \cmidrule(lr){8-11} \cmidrule(lr){12-14} \cmidrule(lr){15-16} \cmidrule(lr){17-18} \cmidrule(lr){19-20}
       & 0.5B & 1.5B & 3B & 7B & 14B & 32B & 1B & 3B & 8B & 8B-M & 7B & 7B & 7B & 3.8B & 14B & 1B & 32B & Avg & Sig.\\
      \midrule
      A / B / C / D & 9.98 & 14.59 & -5.39 & -32.52 & -3.74 & -6.28 & -0.91 & -9.33 & -9.60 & -17.31 & -1.59 & 1.53 & -2.21 & 0.24 & 0.88 & 0.52 & -36.00 & -5.71 & No \\
      1 / 2 / 3 / 4 & 2.87 & 0.10 & 0.10 & -1.21 & -1.28 & -1.75 & -3.03 & -2.06 & -2.29 & -1.09 & -0.66 & -4.48 & -2.13 & -7.44 & -6.63 & 0.00 & -0.82 & -1.87 & Yes \\
      I / II / III / IV & -0.08 & 2.23 & -19.96 & -41.64 & -25.94 & -26.20 & -2.38 & -14.69 & -1.51 & -15.82 & 0.95 & 1.93 & -6.08 & 0.31 & 0.47 & -1.90 & -33.97 & -10.84 & Yes\\
      Min-Max Range & 2.95 & -2.13 & 11.93 & 40.43 & 24.34 & 24.11 & 1.47 & 12.63 & -0.78 & 14.73 & -1.61 & -6.41 & 3.95 & -7.71 & -7.10 & 1.99 & 35.18 & 8.97 & -\\
      \midrule
      A / B / C / D & 1.14 & 7.51 & -1.65 & -27.42 & -5.34 & -5.34 & -1.68 & -3.21 & 7.63 & -6.52 & 0.29 & 2.86 & -3.00 & 0.16 & 0.12 & 1.85 & -17.27 & -2.93 & Yes\\
      1 / 2 / 3 / 4 & 0.68 & -1.21 & -2.67 & -2.10 & -3.38 & -4.68 & -3.84 & -3.90 & -2.46 & -2.18 & -1.79 & -3.30 & -2.90 & -5.97 & -0.85 & 0.27 & -2.99 & -2.55 & Yes\\
      I / II / III / IV & 0.07 & -0.06 & -8.32 & -29.43 & -30.40 & -24.58 & -3.60 & -9.74 & -2.00 & -6.85 & 1.12 & -2.59 & -6.53 & 0.05 & 0.12 & -3.28 & -18.60 & -8.51 & Yes\\
      Min-Max Range & 0.61 & -1.15 & 0.73 & 27.33 & 25.06 & 19.24 & 1.92 & 5.84 & -0.46 & 4.67 & -2.91 & -0.71 & 3.63 & -6.02 & -0.97 & 3.55 & 15.61 & 5.96 & -\\
      \bottomrule
    \end{tabular}%
  }
  \caption{\textbf{Performance Change After Instruction Removal.}
Acc = (no-instruction accuracy) $-$ (with-instruction accuracy). 
Range expansion = (no-instruction range) $-$ (with-instruction range). 
Negative Acc indicates degradation; positive range values indicate increased formatting sensitivity. 
Five of six label-format groups show statistically significant degradation.}

  \label{tab:main_results_no_instruct}
\end{table*}

\begin{table*}[t]
  \centering
  \resizebox{\textwidth}{!}{%
    \begin{tabular}{l*{19}{c}}
      \toprule
      \multirow{2}{*}{Option Labels} & \multicolumn{6}{c}{QWEN} & \multicolumn{4}{c}{LLAMA} & \multicolumn{3}{c}{MISTRAL} & \multicolumn{2}{c}{PHI 3} & \multicolumn{2}{c}{OLMo-2} & \multicolumn{2}{c}{STATS}\\
      \cmidrule(lr){2-7} \cmidrule(lr){8-11} \cmidrule(lr){12-14} \cmidrule(lr){15-16} \cmidrule(lr){17-18} \cmidrule(lr){19-20}
       & 0.5B & 1.5B & 3B & 7B & 14B & 32B & 1B & 3B & 8B & 8B-M & 7B & 7B & 7B & 3.8B & 14B & 1B & 32B & Avg & Std.\\
      \midrule
      A / B / C / D & 15.17 & 7.75 & 6.62 & 22.99 & 17.17 & 23.31 & 24.89 & 23.43 & 7.29 & 21.98 & 0.44 & 0.39 & 20.89  & 0.00 & 0.00 & 23.26 & 20.78 & 13.90 & 9.77\\
      1 / 2 / 3 / 4 & 25.00 & 24.86 & 26.13 & 26.02 & 26.44 & 26.73 & 25.28 & 23.84 & 25.55 & 23.88 & 24.90 & 23.16 & 25.07 & 14.23 & 5.56 & 25.13 & 25.06 & 23.34 & 5.37\\
      I / II / III / IV & 0.67 & 0.26 & 9.33 & 25.33 & 23.38 & 23.69 & 24.80 & 23.56 & 6.89 & 17.21 & 0.27 & 0.59 & 14.38 & 0.00 & 0.00 & 24.49 & 25.62 & 12.97 & 11.05\\
      Min-Max Range & 24.33 & 24.60 & 19.51 & 3.03 & 9.27 & 3.42 & 0.48 & 0.41 & 18.66 & 6.67 & 24.63 & 22.77 & 10.69 & 14.23 & 5.56 & 1.87 & 4.84 & 10.37 & 9.16\\
      \midrule
      A / B / C / D & 3.84 & 0.38 & 0.75 & 7.21 & 5.66 & 8.28 & 11.33 & 9.47 & 1.62 & 9.18 & 0.10 & 0.35 & 7.10 & 0.00 & 0.00 & 8.91 & 9.84 & 4.94 & 4.22\\
      1 / 2 / 3 / 4 & 11.37 & 11.09 & 11.52 & 10.95 & 10.95 & 11.22 & 11.76 & 11.03 & 11.78 & 11.32 & 11.07 & 10.12 & 11.38 & 4.36 & 3.16 & 9.90 & 10.99 & 10.23 & 2.49\\
      I / II / III / IV & 0.03 & 0.10 & 1.34 & 10.40 & 9.56 & 9.77 & 9.20 & 10.27 & 4.26 & 6.89 & 0.10 & 0.24 & 5.18 & 0.00 & 0.00 & 10.08 & 11.12 & 5.21 & 4.63\\
      Min-Max Range & 11.34 & 10.99 & 10.77 & 3.74 & 5.29 & 2.94 & 2.56 & 1.56 & 10.16 & 4.43 & 10.97 & 9.88 & 6.20 & 4.36 & 3.16 & 1.17 & 1.28 & 5.29 & 3.86\\
      \bottomrule
    \end{tabular}%
  }
  \caption{Model accuracy on MMLU (top) and MMLU-Pro (bottom). Each row shows the mean accuracy across models for a given instruction format. The final row for each dataset reports the min–max accuracy range.}
  \label{tab:main_results_random_baseline}
\end{table*}

\begin{table*}[t]
  \centering
  \resizebox{\textwidth}{!}{%
    \begin{tabular}{l*{22}{c}}
      \toprule
      \multirow{2}{*}{Option Labels} & \multicolumn{7}{c}{QWEN} & \multicolumn{6}{c}{LLAMA} & \multicolumn{3}{c}{MISTRAL} & \multicolumn{2}{c}{PHI 3} & \multicolumn{2}{c}{OLMo-2} & \multicolumn{2}{c}{STATS}\\
      \cmidrule(lr){2-8} \cmidrule(lr){9-14} \cmidrule(lr){15-17} \cmidrule(lr){18-19} \cmidrule(lr){20-21} \cmidrule(lr){22-23}
       & 0.5B & 1.5B & 3B & 7B & 14B & 32B & 72B & 1B & 3B & 8B & 8B-M & 70B & 70B & 7B & 7B & 7B & 3.8B & 14B & 1B & 32B & Avg & Sig.\\
      \midrule
      A / B / C / D & -0.74 & -0.72 & 0.88 & -2.53 & -5.90 & -0.63 & 6.39 & -4.69 & -6.14 & 26.56 & 0.94 & 15.75 & 26.65 & -1.16 & 6.75 & -1.67 & 0.01 & 0.00 & 2.92 & 0.20 & 3.14 & No\\
      1 / 2 / 3 / 4 & 1.32 & 5.37 & 9.65 & 0.26 & -0.54 & 0.40 & 0.91 & -2.87 & -4.44 & -13.62 & 0.50 & -5.56 & -1.14 & 0.75 & -0.12 & -1.42 & 1.63 & 2.83 & 1.42 & 0.52 & -0.21 & No\\
      I / II / III / IV & 0.11 & 1.15 & 4.12 & -1.95 & 3.45 & 2.26 & 6.06 & -1.77 & -8.30 & 9.77 & 15.15 & 34.97 & -10.06 & 1.18 & 7.88 & -0.34 & 0.01 & 0.00 & -0.65 & 3.26 & 3.32 & No \\
      Min-Max Range & 1.21 & 4.22 & 8.77 & 2.21 & -6.11 & -2.47 & -0.72 & -1.60 & 3.86 & -14.21 & -10.29 & -38.53 & -28.51 & -0.43 & -8.00 & -1.08 & 1.62 & 2.83 & 2.07 & -1.24 & -4.32 & -\\
      \midrule
      Is 3-shot Mean higher? & \cmark & \cmark & \cmark & \xmark & \xmark & \cmark & \cmark & \xmark & \xmark & \cmark & \cmark & \cmark & \cmark & \cmark & \cmark & \xmark & \cmark & \cmark & \cmark & \cmark & - & - \\
      Is 3-shot Range smaller? & \xmark & \xmark & \xmark & \xmark & \cmark & \cmark & \cmark & \cmark & \xmark & \cmark & \cmark & \cmark & \cmark & \cmark & \cmark & \cmark & \xmark & \xmark & \xmark & \cmark & - & - \\
      \bottomrule
    \end{tabular}%
  }
  \caption{Impact of 3-shot in-context examples across models and label formats on MMLU-Pro. Positive values in the first three result rows indicate improvement in accuracy over 0-shot performance, while negative values in the `Min-Max Range` row indicate a reduction in performance variability across different label formats (smaller is better). The second-to-last row (\cmark/\xmark) shows whether the 3-shot mean accuracy exceeds the original mean, and the last row (\cmark/\xmark) indicates whether the range decreased, reflecting more consistent instruction adherence.}
  \label{tab:main_results_3shot}
\end{table*}

\subsection{Model Selection and Decoding} We evaluate 25 LLMs (20 of which are instruction-tuned), spanning 0.5B to 72B parameters. All models were quantized to 4-bit precision to minimize memory usage and enable deployment on a single NVIDIA RTX 5090 GPU. The only exceptions were the 70B-class models, which required CPU offloading due to their size. And for these 70B models, we report results using only 5\% of samples from each category in each dataset (i.e., around 700 and 600 samples for MMLU and MMLU-Pro, respectively. Table \ref{tab:models} in Appendix~\ref{sec:appendix_models_used} summarizes the models used in our experiments. All experiments use 0-shot greedy decoding to eliminate the effect of randomness.


\subsection{Datasets} MCQ datasets offer a controlled environment \cite{robinson2023leveraginglargelanguagemodels} where tasks are well-defined, outputs are constrained, and variations in instructions or formats can be cleanly isolated, making them ideal for studying instruction-following capability of \textbf{simple atomic instructions}. We pick two benchmarks: MMLU \cite{hendrycks2021measuringmassivemultitasklanguage} and MMLU-PRO \cite{wang2024mmluprorobustchallengingmultitask}. The MMLU dataset covers 57 subjects across diverse domains, designed to evaluate models’ general knowledge and reasoning abilities. MMLU-PRO extends MMLU by introducing more challenging questions and increasing the number of answer options from four to ten. We modify these datasets by adding explicit instructions, formatting them in the form of instruction–question–options.

\subsection{Prompt Setups}
To rigorously assess atomic instruction-following capability, defined as the model’s ability to execute basic, self-contained directives, we conduct three controlled experiments (Figure 1). \textbf{Experiment 1} tests explicit instruction adherence by presenting identical task content with three distinct label formats: Alphabetic ("Pick the correct option (A,B,C,D). Respond with only the letter"), Numeric ("Pick the correct option (1,2,3,4). Respond with only the number"), and Roman ("Pick the correct option (I,II,III,IV). Respond with only the Roman numeral"). Crucially, semantic content remains constant across formats.

\textbf{Experiment 2} removes instructional preambles entirely while preserving all other elements (question stems and option content) across the same three label formats. This quantifies the extent to which instructions mitigate formatting biases. \textbf{Experiment 3} deprives models of option content, presenting only labels (e.g., "A. ", "1. ", "I. ") alongside instructions and questions. This tests whether models can execute directives based solely on symbolic placeholders.

\textbf{Experiment 4} To investigate whether failures in atomic instruction following can be mitigated through explicit guidance, we extend our primary experimental setup by incorporating in-context 3-shot examples during inference. Unlike standard 0-shot prompts (Experiment 1), we prepend three exemplar question-answer pairs that explicitly demonstrate the required response format, while keeping the semantic content of the target question unchanged. This design tests whether providing clear, concrete examples can guide the model to generate outputs that strictly adhere to the instructed format (e.g., "A" instead of "1"), directly addressing the failure modes observed in prior experiments.

\subsection{Evaluation Metrics} We adopt a strict evaluation protocol to assess models' ability to follow instructions and select correct answers. Specifically, a model's response is considered correct only if it \textbf{exactly matches} the symbol corresponding to the correct answer option, as explicitly mentioned in the prompt. This stringent approach avoids reliance on auxiliary classifiers or probability-based methods \cite{brown2020languagemodelsfewshotlearners, shi2024thoroughexaminationdecodingmethods}, and instead directly tests whether the model can adhere to the prompt format, and essential capability for real-world instruction-following tasks. At the question level, accuracy is 1 for an exact match with the correct option symbol and 0 otherwise. For each benchmark, per-subsection accuracies are first averaged, and the final benchmark accuracy is obtained by averaging across subsections.

\subsection{Analysis of Generated Answers} Beyond accuracy, we perform a fine-grained analysis of model outputs to measure label fidelity - the proportion of generated responses that fall within the acceptable set of labels (e.g., {A, B, C, D}). This analysis, conducted for both 0-shot and 3-shot settings, assesses whether models can reliably produce syntactically valid outputs as instructed, irrespective of their factual correctness.

\section{Results and Analysis}

\subsection{Main Results: Atomic Instruction Failure in Instruction-Tuned LLMs}

\textbf{Overall Performance \& Statistical Significance.} With reference to Table~\ref{tab:main_results}, Numerically-labeled prompts achieved the highest mean accuracy (MMLU: 63.06\%; MMLU-Pro: 35.50\%) with zero failures. In stark contrast, semantically equivalent alphabetic and Roman formats showed significantly lower accuracy (alphabetic: 42.16\% MMLU, 22.33\% MMLU-Pro; Roman: 32.62\% MMLU, 20.17\% MMLU-Pro) and high failure rates (5-10 models below random baseline). Post-hoc Wilcoxon tests with Bonferroni correction confirmed all pairwise differences are statistically significant ($p < 0.05$), validating that surface-level formatting, not semantic content, drives performance variation.

\textbf{Model-Specific Trends.} We observe three key patterns: First, while larger models like Qwen-72B achieved higher peak accuracy (84.06\%), they exhibited wider performance ranges across formats (up to 7.81 percentage points [pp]) than smaller counterparts, indicating that scale does not eliminate formatting sensitivity and may even exacerbate it. Second, Phi-3 models consistently failed on alternative formats regardless of size (near 0\% accuracy), demonstrating that instruction adherence is not size-dependent. Third, Mistral showed consistent gains across formats from V0.1 to V0.3 (e.g., +15-30pp on Roman labels), proving instruction-following can be improved without scaling. However, Mistral V0.3 still lags behind same-size Qwen models in both peak accuracy and robustness, highlighting the need to measure both task performance and instruction-following capability.

\textbf{Core Failure Analysis.} The extreme performance divergence, up to 72.20pp range for LLaMA-70B—between semantically identical formats reveals that IT-LLMs fundamentally fail to execute atomic instructions. This is particularly striking for Roman, where 10+ models performed below random chance despite explicit task directives. We attribute this to \textbf{implicit formatting priors} in instruction tuning: models develop bias toward numerically-structured outputs, undermining their ability to treat semantically equivalent labels as interchangeable placeholders. This demonstrated failure to execute atomic instructions despite explicit directives directly exposes fundamental limitations in how IT-LLMs process simple commands. 

Having demonstrated IT-LLMs' fragility to prompt variations, we now investigate whether instructions actually confer performance benefits. By removing instructions while keeping all other factors constant, we evaluate performance differences across option-label styles, revealing that instructions provide only partial mitigation of formatting biases.

\subsection{Instruction-Tuning Benefits and Limitations}

Removing task instructions significantly degraded performance in five of six label-format conditions (Wilcoxon tests, Bonferroni-corrected $\alpha = 0.0167$), confirming their critical role in knowledge-intensive tasks. As shown in Table~\ref{tab:main_results_no_instruct}:

On MMLU, removing instructions caused statistically significant performance degradation for Numeric labels (W = 15, adjusted $p = 0.018$) and Roman labels (W = 24, adjusted $p = 0.033$), but not for Alphabetic labels (W = 38, adjusted $p = 0.214$).

On MMLU-Pro, instruction removal significantly harmed performance across all label formats: Alphabetic (W = 0, adjusted $p = 0.001$), Numeric (W = 0, adjusted $p < 0.001$), and Roman (W = 2, adjusted $p = 0.003$).

Performance degradation was most severe for Roman (MMLU: -10.84\%; MMLU-Pro: -8.51\%) compared to alphabetic/numeric formats. Crucially, the \textit{absence of instruction} amplified variability between semantically equivalent label formats, with average min-max range expansions of 8.97\% (MMLU) and 5.96\% (MMLU-Pro). These results underscore the benefits of providing instructions, in terms of both task performance and consistency among prompt variations.

Nevertheless, we note that even with explicit instructions, a persistent performance gap between label formats remains. Even after instruction-tuning, LLMs fail to execute simple atomic instructions - specifically, the unambiguous directive to treat option labels as equivalent semantic placeholders. This irreducible symbolic bias demonstrates fundamental limitations in overcoming symbolic reasoning deficiencies through instruction tuning, exposing unexpected fragility to \textbf{basic} directives and necessitating careful prompt design for instruction-dependent tasks.

\subsection{A Fundamental Failure: Instruction Following Without Content}

Our content-deprived instruction following analysis (Table~\ref{tab:main_results_random_baseline}) reveals a fundamental fragility in IT-LLMs. When option content is removed, leaving only labels with instruction-question pairs, models fail to reliably execute the given instruction. Performance approaches random baselines exclusively with numerical labels, collapsing with semantically identical alternatives. MMLU (25\% baseline), numerical labels yield chance-level accuracy (mean=23.34\%, $p=0.7960$), while alphabetic (mean=13.90\%, $p<0.001$) and Roman (mean=12.97\%, $p<0.001$) formats show significant degradation. Similarly on MMLU-Pro (10\% baseline), numerical labels show marginal improvement (mean=10.23\%, $p=0.0442$), whereas alphabetic (mean=4.94\%, $p<0.001$) and Roman (mean=5.21\%, $p<0.01$) formats perform below chance. These results lend further support to our claim that IT-LLMs cannot execute simple atomic instructions.

\subsection{The Ineffectiveness of Few-Shot Learning}

We evaluate the impact of 3-shot in-context guidance on atomic instruction-following across multiple models and label formats. Table~\ref{tab:main_results_3shot} summarizes the change in accuracy relative to 0-shot performance (Experiment 1) and the variability across label formats.

Accuracy Changes. The first three rows show the changes in accuracy under the 3-shot setting. Positive values indicate that providing explicit 3-shot examples boosts performance. Across the 20 models, 15 demonstrate gains, highlighting that explicit examples can help in certain cases. However, the improvements are inconsistent across models and formats.

Consistency Across Label Formats. The Min-Max Range row captures the spread of performance across different label schemes, where negative values indicate reduced variability. Only 12 out of 20 models show a decrease in range. This result suggests that 3-shot guidance only partially mitigates format inconsistency. For a significant portion of models, adherence to an instructed output format remains a brittle capability, not robustly solved by providing a few exemplars.

To confirm these observations statistically, we performed a Wilcoxon signed-rank test comparing 0-shot and 3-shot performance across models for each label format. The results indicate \textbf{no statistically significant improvements for any format}: for Alphabetic labels, the Wilcoxon statistic is 95.0 (uncorrected p = 1.0, Bonferroni-corrected p = 1.0); for Numeric labels, the statistic is 102.0 (uncorrected p = 0.927, corrected p = 1.0); and for Roman labels, the statistic is 72.0 (uncorrected p = 0.355, corrected p = 1.0). These results confirm that incorporating 3-shot explicit examples does not yield statistically significant improvements.

Taken together, the descriptive and statistical analyses demonstrate that, even with explicit examples, IT-LLMs consistently fail to follow semantically trivial instructions when superficial features such as option label formats are varied. This reinforces our key finding that atomic instruction-following remains a challenging and largely unsolved problem.






\section{Analysis of Generated Answers: Label Fidelity and Bias}

\begin{table*}[t]
  \centering
  \resizebox{\textwidth}{!}{%
    \begin{tabular}{l*{22}{c}}
      \toprule
      \multirow{2}{*}{Option Labels} & \multicolumn{7}{c}{QWEN} & \multicolumn{6}{c}{LLAMA} & \multicolumn{3}{c}{MISTRAL} & \multicolumn{2}{c}{PHI 3} & \multicolumn{2}{c}{OLMo-2} & \multicolumn{2}{c}{STATS}\\
      \cmidrule(lr){2-8} \cmidrule(lr){9-14} \cmidrule(lr){15-17} \cmidrule(lr){18-19} \cmidrule(lr){20-21} \cmidrule(lr){22-23}
       & 0.5B & 1.5B & 3B & 7B & 14B & 32B & 72B & 1B & 3B & 8B & 8B-M & 70B & 70B & 7B & 7B & 7B & 3.8B & 14B & 1B & 32B & Avg & Sig.\\
      \midrule
      A / B / C / D & 5.98 & 14.06 & 7.07 & 84.73 & 97.00 & 94.58 & 85.19 & 88.77 & 90.21 & 15.00 & 68.80 & 32.15 & 16.16 & 24.78 & 34.43 & 69.30 & 0.01 & 0.00 & 99.90 & 96.27 & 51.22 & -\\
      1 / 2 / 3 / 4 & 98.02 & 99.58 & 99.96 & 99.91 & 99.98 & 99.99 & 100.00 & 99.15 & 98.35 & 99.83 & 98.15 & 100.00 & 99.83 & 99.80 & 99.29 & 99.98 & 97.11 & 87.77 & 99.92 & 99.60 & 98.81 & -\\
      I / II / III / IV & 0.61 & 6.53 & 18.53 & 99.31 & 88.22 & 88.84 & 69.02 & 77.79 & 83.70 & 9.29 & 46.97 & 10.61 & 16.50 & 2.90 & 20.24 & 58.05 & 0.02 & 0.01 & 85.29 & 99.94 & 44.12 & -\\
      Mean & 34.87 & 40.06 & 41.85 & 94.65 & 95.07 & 94.47 & 84.74 & 88.57 & 90.75 & 41.37 & 71.31 & 47.59 & 44.16 & 42.49 & 51.32 & 75.78 & 32.38 & 29.26 & 94.04 & 99.60  & 64.72 & -\\
      Min-Max Range & 97.41 & 93.05 & 92.89 & 15.18 & 11.76 & 11.15 & 30.98 & 21.36 & 14.65 & 90.54 & 51.18 & 89.39 & 83.67 & 96.90 & 79.05 & 41.93 & 97.10 & 87.77 & 14.63 & 0.67 & 54.69 & -\\
      \midrule
      A / B / C / D & 0.55 & 14.16 & 10.04 & 74.37 & 80.09 & 89.81 & 97.64 & 80.17 & 85.73 & 79.10 & 73.80 & 67.85 & 71.04 & 24.59 & 56.26 & 65.85 & 0.03 & 0.01 & 94.22& 98.10 & 58.17 & No\\
      1 / 2 / 3 / 4 & 98.50 & 100.00 & 99.98 & 99.93 & 99.99 & 99.98 & 100.00 & 99.98 & 99.46 & 99.73 & 99.60 & 99.83 & 100.00 & 99.48 & 99.37 & 99.91 & 99.81 & 92.14 & 99.99 & 99.84 & 99.38 & No\\
      I / II / III / IV & 0.73 & 13.06 & 35.32 & 82.40 & 94.88 & 96.47 & 83.00 & 72.09 & 70.79 & 65.58 & 54.60 & 83.16 & 89.73 & 4.65 & 40.77 & 57.88 & 0.01 & 0.00 & 78.41 & 99.24 & 56.14 & No \\
      Mean & 33.26 & 42.41 & 48.45 & 85.57 & 91.65 & 95.42 & 93.55 & 84.08 & 85.33 & 81.47 & 76.00 & 83.61 & 85.52 & 42.91 & 65.47 & 74.55 & 33.28 & 30.72 & 90.87 & 99.06 & 71.16 & -\\
      Min-Max Range & 97.95 & 86.94 & 89.94 & 25.56 & 19.90 & 10.17 & 17.00 & 27.89 & 28.67 & 34.15 & 45.00 & 31.98 & 28.96 & 94.83 & 58.60 & 42.03 & 99.80 & 92.14 & 21.58 & 1.74 & 45.01 & -\\
      \midrule
      Is 3-shot Mean higher? & \xmark & \cmark & \cmark & \xmark & \xmark & \cmark & \cmark & \xmark & \xmark & \cmark & \cmark & \cmark & \cmark & \cmark & \cmark & \xmark & \cmark & \cmark & \xmark & \xmark & - & - \\
      Is 3-shot Range smaller? & \xmark & \cmark & \cmark & \xmark & \xmark & \cmark & \cmark & \xmark & \xmark & \cmark & \cmark & \cmark & \cmark & \cmark & \cmark & \xmark & \xmark & \xmark & \xmark & \xmark & - & - \\

      \bottomrule
    \end{tabular}%
  }
  \caption{Proportion of generated labels that are within the predefined acceptable set across label styles and models on MMLU-Pro. The top table reports results for the 0-shot setting, and the bottom for the 3-shot setting. While the average proportion of acceptable generations increases slightly in the 3-shot setting, the difference is not statistically significant. The Min–Max range also decreases but remains high at 45.01\%.}
  \label{tab:main_results_output_analysis}
\end{table*}

To assess how reliably models adhere to discrete answer formats, we measure the proportion of generated outputs that fall within a predefined acceptable set of labels (e.g., “A” from {A, B, …, J}). Table \ref{tab:main_results_output_analysis} summarizes these proportions across label styles, model families, and scales, under both 0-shot (top) and 3-shot (bottom) prompting conditions. This evaluation provides a fine-grained view of atomic instruction-following - whether a model can produce syntactically valid responses with or without explicit examples. Overall, we observe that while the inclusion of three in-context examples leads to slightly higher rates of acceptable label generation, the improvement is not statistically significant. Furthermore, substantial variability remains across models and label formats, revealing that in-context learning alone is insufficient to eliminate the underlying response biases- learned during either pretraining or instruction-tuning.

\subsection{Label Fidelity}

 Across all models and label formats, introducing three in-context demonstrations leads to only modest increases in the proportion of acceptable outputs. Statistical analysis using pairwise Wilcoxon signed-rank tests with Bonferroni correction confirms that these improvements are not significant (all adjusted p > 0.05). In other words, providing three examples per prompt is insufficient to reliably enhance atomic instruction-following, as the underlying biases of these models continue to shape their generation behavior.

While the 3-shot setting slightly narrows the Min--Max range of acceptable outputs (from 54.69\% to 45.01\%), considerable variability persists across label styles. For instance, most models handle numeric labels robustly (nearly 100\% adherence), yet struggle markedly with letter-based or Roman labels - a gap that exceeds 80 percentage points in some cases. Even with demonstrations, models frequently produce outputs outside the acceptable set, underscoring the difficulty of achieving consistent, discrete label generation. These results suggest that few-shot prompting alone cannot fully overcome the intrinsic limitations of models in adhering to strict label constraints or normalizing across output styles.

\subsection{Error Analysis: Preference for Numeral Responses}

Referring to Table~\ref{tab:generated_output_numeral} in Appendix~\ref{sec:generated_output_numeral}, we observe that in both the alphabetical and Roman setups, many models continue to produce numeral responses despite explicit instructions to use the corresponding label format. When numeral outputs are also considered acceptable - alongside the intended label style - the number of models whose proportion of acceptable responses exceeds 80\% increases markedly: from 7 to 13 under the 0-shot setting, and from 6 to 15 under the 3-shot setting. This indicates a strong bias toward numeral-style responses, even when the prompt explicitly specifies otherwise.

\subsection{Effects of Model Family and Scale}

Beyond overall averages, examining performance across model families and scales reveals systematic patterns. Within the Qwen familiy, smaller models ($\leq$3B parameters) frequently generate out-of-set labels, whereas mid- to large-scale models (>7B) consistently produce high proportions of acceptable outputs, often exceeding 90\%. For instance, Qwen’s 3B model achieves only 7.07\% adherence on letter-based labels, whereas the 7B variant jumps to 84.73\%. This indicates that model scale strongly influences the fidelity of discrete label generation.

By contrast, the LLaMA family exhibits a more nuanced pattern: while the 70B model performs nearly perfectly on numeric labels, it fails catastrophically on symbolic ones such as Roman (e.g., the 3.1 variant only got 10.61\%). Interestingly, smaller LLaMA variants (1B and 3B) achieve much higher overall averages (around 90\%), suggesting that scaling alone does not guarantee uniform improvements in output fidelity.

Mistral exhibits highly variable adherence, performing well on numeric labels but inconsistently on letters and Roman. Similarly, Phi-3 and OLMo-2 models achieve near-perfect adherence for numeric formats but lower accuracy for symbolic labels, reflecting a potential sensitivity to label type and tokenization. While these family-specific differences likely stem from differences in pretraining data composition and instruction-tuning strategies, distinguishing between these sources remains an open question for future work.

Taken together, these findings highlight that few-shot prompting can modestly improve adherence to expected label formats, but the improvement is neither robust nor uniform across architectures, scales, or label types.

\subsection{Ability to Learn from Examples}

To examine how effectively models leverage demonstrations, we compare the change in acceptable-label proportions and Min--Max ranges between the 0-shot and 3-shot settings. Out of 20 evaluated models, 12 exhibit an increase in the proportion of acceptable outputs, while 10 show a decrease in variability (smaller Min--Max range). These mixed outcomes indicate that not all models are equally capable of internalizing the mapping between input and desired discrete output from a few examples.

Interestingly, this ability does not scale monotonically with model size. Within the Qwen family, for example, only the 1.5B, 3B, 32B, and 72B variants benefit from demonstrations, whereas the 0.5B, 7B, and 14B models fail to leverage them effectively. This suggests that the mechanisms supporting in-context learning are not solely a function of SIZE, but may depend on factors such as pretraining dynamics, instruction-tuning objectives, or decoder stability. Overall, while some models exhibit partial few-shot adaptability, the improvements remain small and inconsistent, reinforcing the conclusion that in-context learning alone is insufficient to drive robust gains in atomic instruction-following.

\section{Ablation: Effect of Instruction Tuning vs. Base Models}

\begin{table}[t]
  \centering
  \resizebox{\columnwidth}{!}{%
    \begin{tabular}{lccccc}
      \toprule
      \multirow{2}{*}{Option Labels} & \multicolumn{3}{c}{QWEN} & \multirow{2}{*}{LLAMA 8B} & \multirow{2}{*}{MISTRAL 7B}\\
      \cmidrule(lr){2-4}
       & 0.5B & 1.5B & 14B & & \\
      \midrule
      A / B / C / D & 0.65 & -0.04 & 13.93 & 4.61 & 8.01 \\
      1 / 2 / 3 / 4 & 0.83 & 4.82 & 0.96 & 8.13 & -0.06 \\
      I / II / III / IV & 0.03 & 0.34 & 30.34 & 2.02 & 6.76 \\
      Min-Max Range & 0.80 & 4.48 & -25.62 & 6.11 & -6.82 \\
      Is Mean higher? & \cmark & \cmark & \cmark & \cmark & \cmark \\
      Is Range smaller? & \xmark & \xmark & \cmark & \xmark & \cmark \\
      \midrule
      A / B / C / D & 0.17 & 2.23 & 26.98 & 13.60 & -1.33 \\
      1 / 2 / 3 / 4 & 0.42 & 8.02 & 0.67 & -9.16 & -1.64 \\
      I / II / III / IV & 0.14 & 1.12 & 16.64 & 9.99 & -0.55 \\
      Min-Max Range & 0.28 & 6.90 & -26.29 & -9.97 & -1.09 \\
      Is Mean higher? & \cmark & \cmark & \cmark & \cmark & \xmark \\
      Is Range smaller? & \xmark & \xmark & \cmark & \cmark & \cmark \\
      \bottomrule
    \end{tabular}%
  }
  \caption{Acc deltas between instruction-tuned and non-instruction-tuned LLMs on 0-shot (top) and 3-shot (bottom) for MMLU-Pro. Positive values indicate performance improvements after instruction tuning. For the “Min–Max Range” rows, negative values indicate reduced variability across option label styles, reflecting more consistent behavior. The final two rows per dataset denote whether the instruction-tuned model’s mean accuracy increased and whether variability decreased.}
  \label{tab:ablation_non_inst}
\end{table}

To evaluate whether the observed label sensitivity arises from instruction tuning or from pretraining, we compared each instruction-tuned model with its corresponding non-instruction-tuned (base) version across multiple option label formats. Table~\ref{tab:ablation_non_inst} reports the delta performance for MMLU and MMLU-Pro. Positive values indicate that instruction tuning improved accuracy.

Across models, instruction tuning generally improved mean accuracy (9 out of 10 instances). However, the Min–Max Range rows reveal that tuning does not always decreases sensitivity to label formats (only 5 out 10 instances). \textbf{This pattern suggests that instruction tuning enhances task performance but does not uniformly mitigate the label-style sensitivity observed in the base models.}

\section{Conclusion}

Our findings reveal a critical shortcoming: IT-LLMs often fail to treat atomic instructions as abstract operations, instead relying on surface patterns, for instance, numeric labels consistently outperform Roman or alphabetic equivalents despite semantic equivalence. Although explicit instructions improve accuracy, they do not close the gap between semantically identical prompt formats, a task trivial for humans. Moreover, adding three-shot exemplars offers no statistically significant gains in adherence or robustness. Analysis of generated outputs exposes pervasive label infidelity: models frequently produce responses outside the instructed label set, with few-shot learning providing only marginal mitigation. This fragility persists across scales and model families, indicating a fundamental flaw in current instruction-tuning, stronger task performance does not entail better instruction-following. Our results suggest that robust instruction adherence requires explicit optimization or dedicated tuning, rather than emergent scaling. We present a framework for targeted evaluation of atomic instruction-following and advocate for instruction invariance as a key evaluation dimension alongside accuracy. Ultimately, the practical reliability of IT-LLMs will remain limited until they overcome symbolic biases and treat labels such as “B” and “II” as interchangeable reference tokens. Our work establishes that robust atomic instruction-following is a distinct and crucial capability that current instruction-tuning paradigms fail to instill, presenting a key challenge for building truly reliable language models

\section*{Limitations}

Our experiments focus exclusively on multiple-choice question (MCQ) tasks, which offer a controlled environment for isolating instruction adherence and are sufficient for the objectives of this preliminary study. However, future work should extend this analysis to open-ended and multi-turn tasks to capture the broader spectrum of instruction-following behaviors.
Due to hardware constraints, we employ 4-bit quantized models, and for the 70B model class, experiments are conducted on only 5\% of the dataset.
Furthermore, cost limitations restrict our study to publicly available open-weight models, excluding proprietary systems that have demonstrated stronger instruction-following performance.



\bibliography{custom}

\appendix

\section{Model Names and Versions}
\label{sec:appendix_models_used}

\begin{table}[ht]
\centering
\begin{tabular}{|l|c||l|c|}
\hline
\textbf{Model} & \textbf{Ver.} & \textbf{Model} & \textbf{Ver.} \\
\hline
Llama-1B    & 3.2   & OLMo2-1B      & 0425  \\
Llama-3B    & 3.2   & OLMo2-32B     & 0325     \\
\textbf{Llama-8B}    & 3.1   & Phi-mini      & 3       \\
Llama-8B    & Med42 & Phi-medium    & 3     \\
Llama-70B   & 3.1   & \textbf{Qwen-0.5B}     & 2.5   \\
Llama-70B   & 3.3   & \textbf{Qwen-1.5B}     & 2.5   \\
Mistral-7B  & v0.1  & Qwen-3B       & 2.5   \\
\textbf{Mistral-7B}  & v0.2  & Qwen-7B       & 2.5   \\
Mistral-7B  & v0.3  & \textbf{Qwen-14B}      & 2.5   \\
            &       & Qwen-32B      & 2.5   \\
            &       & Qwen-72B      & 2.5   \\
\hline
\end{tabular}
\caption{Models and their versions used. Bold indicates that its non-instruction-tuned version was used too.}
\label{tab:models}
\end{table}

\section{Generated Output Numeral}
\label{sec:generated_output_numeral}

\begin{table*}[t]
  \centering
  \resizebox{\textwidth}{!}{%
    \begin{tabular}{l*{20}{c}}
      \toprule
      \multirow{2}{*}{Option Labels} & \multicolumn{7}{c}{QWEN} & \multicolumn{6}{c}{LLAMA} & \multicolumn{3}{c}{MISTRAL} & \multicolumn{2}{c}{PHI 3} & \multicolumn{2}{c}{OLMo-2} \\
      \cmidrule(lr){2-8} \cmidrule(lr){9-14} \cmidrule(lr){15-17} \cmidrule(lr){18-19} \cmidrule(lr){20-21} 
      
       & 0.5B & 1.5B & 3B & 7B & 14B & 32B & 72B & 1B & 3B & 8B & 8B-M & 70B & 70B & 7B & 7B & 7B & 3.8B & 14B & 1B & 32B \\
      \midrule
      A / B / C / D & 91.89 & 90.42 & 96.85 & 94.38 & 97.92 & 94.66 & 98.15 & 93.35 & 94.74 & 91.04 & 81.32 & 67.85 & 82.49 & 61.27 & 39.81 & 95.89 & 0.02 & 0.07 & 97.61 & 99.63 \\
      I / II / III / IV & 83.40 & 69.54 & 27.75 & 99.80 & 98.63 & 91.03 & 96.63 & 91.54 & 95.71 & 90.53 & 81.23 & 83.00 & 89.90 & 78.13 & 21.88 & 77.80 & 0.06 & 0.05 & 92.90 & 99.94 \\
      Mean & 87.65 & 79.98 & 62.30 & 97.09 & 98.28 & 92.85 & 97.39 & 92.45 & 95.23 & 90.79 & 81.28 & 75.43 & 86.20 & 69.70 & 30.85 & 86.85 & 0.04 & 0.06 & 95.26 & 99.79 \\
      Is Mean > 80? & \cmark & \xmark & \xmark & \cmark & \cmark & \cmark & \cmark & \cmark & \cmark & \cmark & \cmark & \xmark & \cmark & \xmark & \xmark & \cmark & \xmark & \xmark & \cmark & \cmark \\
      \midrule
      A / B / C / D & 92.77 & 89.34 & 73.16 & 93.28 & 81.45 & 91.61 & 99.66 & 89.08 & 92.65 & 91.40 & 84.01 & 80.81 & 82.49 & 61.80 & 78.20 & 94.93 & 2.04 & 0.03 & 96.73 & 98.87\\
      I / II / III / IV & 90.22 & 75.09 & 40.13 & 97.37 & 96.60 & 99.37 & 97.81 & 89.57 & 92.69 & 89.69 & 80.76 & 91.08 & 93.43 & 58.04 & 53.93 & 82.52 & 7.73 & 0.05 & 96.72 & 99.50 \\
      Mean & 91.50 & 82.22 & 56.65 & 95.33 & 89.03 & 95.49 & 98.74 & 89.33 & 92.67 & 90.55 & 82.39 & 85.95 & 87.96 & 59.92 & 66.07 & 88.73 & 4.89 & 0.04 & 96.73 & 99.19 \\
      Is Mean > 80? & \cmark & \cmark & \xmark & \cmark & \cmark & \cmark & \cmark & \cmark & \cmark & \cmark & \cmark & \cmark & \cmark & \xmark & \xmark & \cmark & \xmark & \xmark & \cmark & \cmark\\
      \bottomrule
    \end{tabular}%
  }
  \caption{MMLU-Pro: Proportion of generated labels that are either numeral or their supposed label styles (either alphabetical or Roman). The top table reports results for the 0-shot setting, and the bottom for the 3-shot setting}
  \label{tab:generated_output_numeral}
\end{table*}

\end{document}